\documentclass[runningheads]{llncs}
\usepackage{makeidx}
\usepackage{graphicx}
\usepackage{amsmath,amssymb} 
\usepackage{color}
\usepackage{comment}

\usepackage{booktabs}

\usepackage{xspace}
\makeatletter
\DeclareRobustCommand\onedot{\futurelet\@let@token\@onedot}
\def\@onedot{\ifx\@let@token.\else.\null\fi\xspace}

\def\ie{\emph{i.e}\onedot} \def\Ie{\emph{I.e}\onedot}

\def\etal{\emph{et al}\onedot}
\makeatother

\begin{document}
	\pagestyle{headings}
	\mainmatter

	\def\GCPR20SubNumber{22}

	\title{Long-Term Anticipation of Activities with Cycle Consistency}

	\titlerunning{Long-Term Anticipation of Activities with Cycle Consistency.}
	\authorrunning{Abu Farha \etal.}
	\author{Yazan Abu Farha$^{1}$ \and Qiuhong Ke$^{2}$ \and Bernt Schiele$^{3}$ \and Juergen Gall$^{1}$}
	\institute{$^{1}$University of Bonn, Germany \\
	           $^{2}$The University of Melbourne, Australia \\
	           $^{3}$MPI Informatics, Germany}

	\maketitle

	\begin{abstract}
With the success of deep learning methods in analyzing activities in videos, 
more attention has recently been focused towards anticipating future activities. 
However, most of the work on anticipation either analyzes a partially observed activity or 
predicts the next action class. 
Recently, new approaches have been 
proposed to extend the prediction horizon up to several minutes in the future and that anticipate 
a sequence of future activities including their durations. While these works decouple the semantic 
interpretation of the observed sequence from the anticipation task, we propose a framework 
for anticipating future activities directly from the features of the observed frames and train it in an end-to-end 
fashion. Furthermore, we introduce a cycle consistency loss over time by predicting the past 
activities given the predicted future. Our framework achieves state-of-the-art results on two 
datasets: the Breakfast dataset and 50Salads.
	\end{abstract}

	\section{Introduction}
	\label{sec:introduction}

\begin{figure}[t]
\begin{center}
\includegraphics[width=\linewidth]{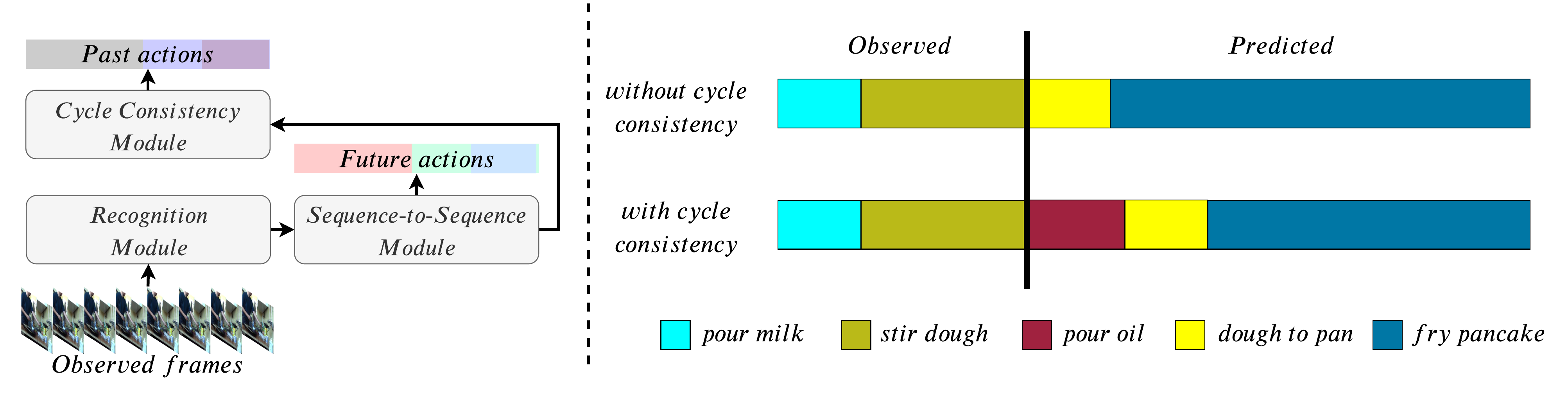}
\end{center}
\vspace{-5mm}
   \caption{(Left) Overview of the proposed approach, which is trained end-to-end and includes a cycle consistency module.
   (Right) Effect of the cycle consistency module. Without the cycle consistency, the network anticipates 
   actions that are plausible based on the previous actions. However, in some cases an essential 
   action is missing. In this example \textit{pour oil}. By using the cycle consistency, we enforce the network to verify if 
   all required actions have been done before. For the action \textit{fry pancake}, \textit{pouring oil} into the pan is 
   required and the cycle consistency resolves this issue. 
   }
\label{fig:new_teaser}
\end{figure}

Humans spend a significant time of their life thinking about the future. Whether thinking about 
their future dream job, or planning for the next research project. Even unconsciously, people tend to anticipate future 
trajectories of moving agents in the surrounding environment and the activities that they will be doing in 
the near future. Such anticipation capability is considered a sign of intelligence and an important factor 
in determining how we interact with the environment and how we make decisions. 

Since anticipation is an important intrinsic capability of human beings, researchers have recently tried 
to model this capability and embed it in intelligent and robotic systems. For example, several approaches 
have been proposed to anticipate future trajectories of pedestrians~\cite{kitani2012activity,alahi2016social}, 
or semantic segmentation of future frames in video~\cite{luc2017predicting,bhattacharyya2018bayesian}. 
These approaches have many applications in autonomous driving and navigation. Another line of 
research focuses on anticipating future activities~\cite{lan2014hierarchical,koppula2016anticipating,vondrick2016anticipating}, which has potential applications in surveillance and human-robot interaction.

While anticipating the next action a few seconds in the future has been addressed in~\cite{vondrick2016anticipating,zeng2017visual,gammulle_2019_predicting}, such short time horizon is insufficient for many applications. 
Service robots, for example, where a robot is continuously interacting with a human, require anticipating a longer time horizon, which rather includes a sequence of future activities 
than only predicting the next action. By anticipating longer in the future, such robots 
can plan ahead and accomplish their tasks more efficiently. Recent approaches, therefore, 
focused on increasing the prediction horizon up to several minutes and predict multiple 
action segments in the future~\cite{abufarha2018when,gammulle2019forecasting,ke2019time}. 
While these approaches successfully predict future activities and their duration, they decouple 
the semantic interpretation of the observed sequence from the anticipation task using a two-step approach. 

Separating the understanding of the past and the anticipation of future has several disadvantages. First, the model is not trained end-to-end, which means that the approach for temporal action segmentation is not optimized for the anticipation task. Second, if there are any mistakes in the temporal action segmentation, these mistakes will be propagated and effect the anticipated activities. Finally, the action labels do not represent all information in the observed video that is relevant for anticipating the 
future. In contrast to these approaches, we propose a sequence-to-sequence model that directly
maps the sequence of observed frames to a sequence of future activities and their duration. 
We then cast the understanding of the past as an auxiliary task by proposing a recognition module, 
which consists of a temporal convolutional neural network and a recognition loss, that is combined 
with the encoder.

Furthermore, as we humans can reason about the past given the future, previous works only aim to predict the future, and it is intuitive that forcing the network to predict the past as well is helpful. To this end, we propose a cycle consistency module that predicts the past activities 
given the predicted future. This module verifies if, for the predicted future actions, all required actions have been done before as illustrated in Fig.~\ref{fig:new_teaser}. In this example, the actions \textit{pour dough to pan} and \textit{fry pancake} are plausible given the previous actions, but the action \textit{pour oil} has been missed. The cycle consistency module then predicts from the anticipated actions, the observed actions. Since \textit{pour dough to pan} and \textit{fry pancake} are the inputs for the cycle consistency 
module, it will predict all the required preceding actions such as \textit{pour milk}, \textit{stir dough} and \textit{pour oil}. However, as \textit{pour oil} is not part of the observations, the cycle consistency module will have a high error, which steers the network to predict \textit{pour oil} in the future actions.


Our contribution is thus three folded: First, we propose an end-to-end model for anticipating a sequence of future activities and their durations. Second, we show that the proposed recognition module improves the sequence-to-sequence model. Third, we propose a cycle consistency module that verifies the predictions.

We evaluate our model on two datasets with untrimmed videos containing many action segments: the 
Breakfast dataset~\cite{kuehne2014language} and 50Salads~\cite{stein2013combining}. Our model is able to 
predict the future activities and their duration accurately achieving superior results compared to the 
state-of-the-art methods. 


\section{Related Work}
\paragraph{Early action detection:} The early action detection task tries to recognize an ongoing activity 
given only a partial observation of that activity. An initial work addressing this task by Ryoo~\cite{ryoo2011human} 
is based on a probabilistic formulation using dynamic bag-of-words of spatio-temporal features. Hoai and 
De la Torre~\cite{hoai2014max} used a max-margin formulation. More recent approaches use recurrent neural 
networks with special loss functions to encourage early activity detection~\cite{ma2016learning,sadegh2017encouraging}. 
In contrast to these approaches, we anticipate future activities without even any partial observations.

\paragraph{Future prediction:} Predicting the future has become one of the main research topics in computer vision. 
Many approaches have been proposed to predict future frames~\cite{mathieu2016deep,liang2017dual}, future human trajectories~\cite{kitani2012activity,alahi2016social}, future human 
poses~\cite{jain2016structural,martinez2017human,ruiz2018human}, 
image semantic segmentation~\cite{luc2017predicting,bhattacharyya2018bayesian}, 
or even full sentences describing future frames or steps in recipes~\cite{sener2018zero,mahmud2019captioning}. 
However, low level representations like pixels of frames or very high level natural language sentences 
cannot be used directly. A complementary research direction is concentrated on anticipating 
activity labels. Lan~\etal~\cite{lan2014hierarchical} predicted future actions using hierarchical representations 
in a max-margin framework. Koppula and Saxena~\cite{koppula2016anticipating} used a spatio-temporal graph 
representation to encode the observed activities and then predict object affordances, trajectories, and 
sub-activities. Instead of directly predicting the future action labels, several approaches were proposed to 
predict future visual representations and then a classifier is trained on top of the predicted representations 
to predict the future labels~\cite{vondrick2016anticipating,gao2017red,zeng2017visual,shi2018action,rodriguez2018action,gammulle_2019_predicting}. 
Predicting future representations has also been used in the literature for unsupervised representations 
learning~\cite{srivastava2015unsupervised}. 
Other approaches use multi-task learning to predict the future activity 
and its starting time~\cite{mahmud2017joint,mehrasa2019variational}, or the future activity and its 
location~\cite{liang_2019_peeking,sun_2019_relational}. Miech~\etal~\cite{miech2019leveraging} modeled the 
transition probabilities between actions and combine it with a predictive model to anticipate future action 
labels. There is a parallel line of research addressing the anticipation task in egocentric 
videos~\cite{furnari2018leveraging,furnari2019would}. However, these approaches are limited 
to very few seconds in the future. In contrast to these approaches, we address the anticipation task for a longer 
time horizon.

Recently, more effort has been dedicated to increase the anticipation horizon. Several methods have 
been proposed to anticipate activities several minutes in the future using RNNs~\cite{abufarha2019uncertainty,abufarha2018when}, temporal convolutions with time-variable~\cite{ke2019time}, or memory 
networks~\cite{gammulle2019forecasting}. While these approaches manage to anticipate activities for 
a longer time horizon, their performance is limited. The approach 
of~\cite{gammulle2019forecasting} relies on the ground-truth action labels of the observations, whereas the 
methods in~\cite{abufarha2019uncertainty,abufarha2018when,ke2019time} follow a two-step approach. \Ie they 
first infer the activities in the observed part, then anticipate the future activities with their corresponding 
duration. These two steps are trained separately, which prevents the model from utilizing the visual cues in 
the observed frames. In contrast to these approaches, our model is trained in one step in an end-to-end fashion.

\paragraph{Cycle consistency:} Cycle consistency has been widely used in computer vision. It was used 
to learn dense correspondence~\cite{zhou2016learning}, image-to-image translation~\cite{zhu2017unpaired}, 
and depth estimation~\cite{godard2017unsupervised}. Recent approaches used cycle consistency in the temporal 
dimension~\cite{dwibedi2019temporal,wang2019learning}. Dwibedi~\etal~\cite{dwibedi2019temporal} introduced an approach 
to learn representations using video alignment as a proxy task and cycle consistency for training. In~\cite{godard2017unsupervised}, 
the appearance consistency between consecutive video frames is used to learn representations that generalize to
different tasks. Motivated by the success of cycle consistency in various applications, we apply it as an additional 
supervisory signal to predict the future activities.


\section{The Anticipation Framework}

\begin{figure*}[t]
\begin{center}
   \includegraphics[width=.97\linewidth]{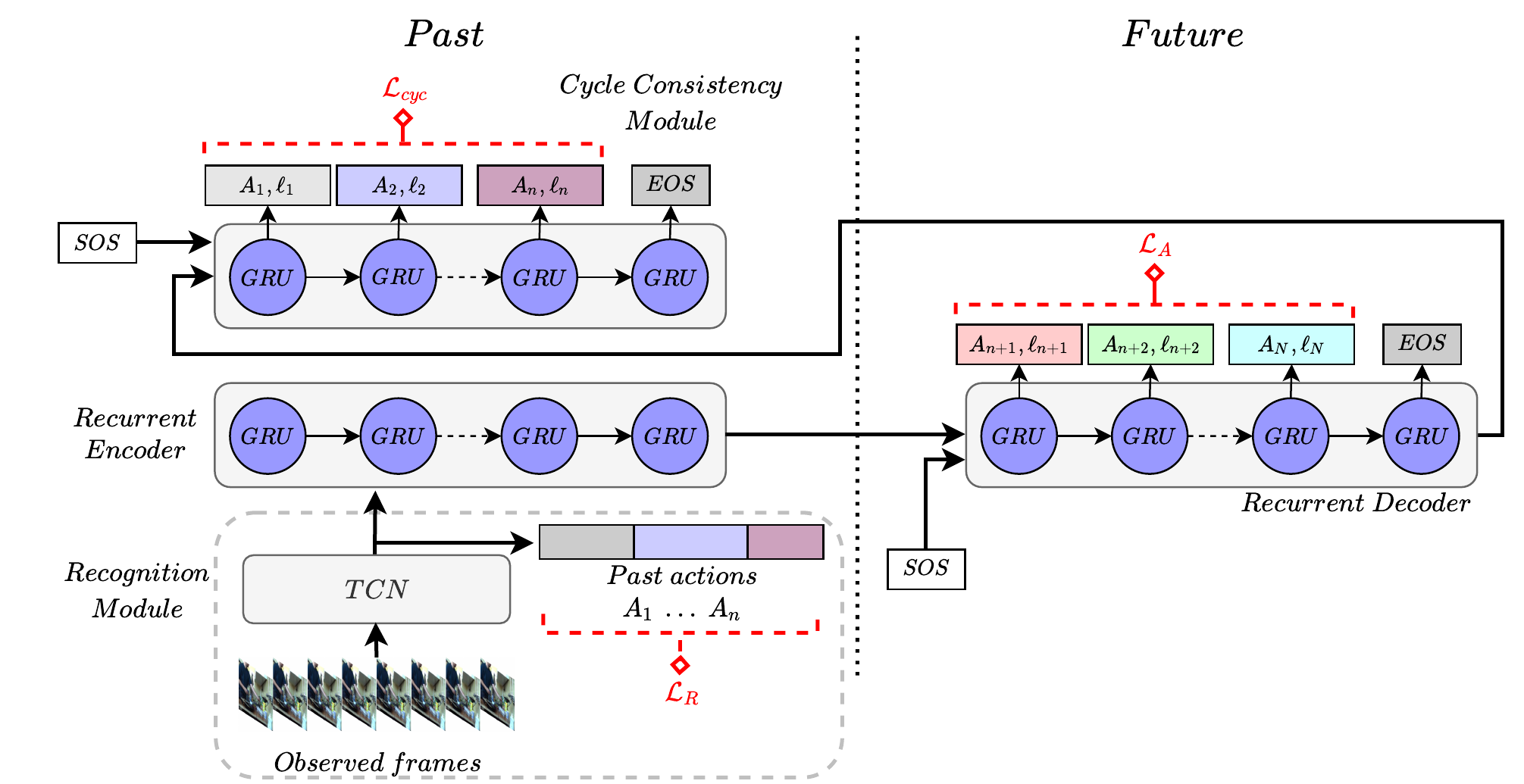}
\end{center}
\vspace{-5mm}
   \caption{Overview of the anticipation framework. The observed frames are passed through 
   a TCN-based recognition module which produces discriminative features for the sequence-to-sequence 
   model. The sequence-to-sequence model predicts the future activities with their duration. In addition, 
   we enforce cycle consistency over time by predicting the past activities given the predicted future.}
\label{fig:model}
\end{figure*}

Given a partially observed video with many activities, we want to predict all the activities 
that will be happening in the remainder of that video with their corresponding duration. Assuming that 
the observed part consists of $t_o$ frames $X_{1:t_o}=(x_1, \dots, x_{t_o})$ corresponding to $n$ activities 
$A_{1:n}=(A_1, \dots, A_n)$, our goal is to predict the future activities $A_{n+1:N}=(A_{n+1}, \dots, A_N)$ and 
their corresponding duration $\ell_{n+1:N}=(\ell_{n+1}, \dots, \ell_N)$, where $N$ is the total number of 
activities in that video. In contrast to the previous approaches that use only the action labels of the 
observed frames for anticipating the future, 
we propose to anticipate the future activities directly from the observed frames. 
First, we propose a sequence-to-sequence model that maps the sequence of features from the observations to a sequence of future 
activities and their duration. Then, we introduce a cycle consistency module that predicts the past activities given the predicted 
future. The motivation of this module is to force the sequence-to-sequence module to encode all the relevant information in the 
observed frames and verify if the predictions are plausible. Finally, we extend the sequence-to-sequence model with a recognition 
module that generates discriminative features that capture the relevant information for anticipating the future. 
An overview of the proposed model is illustrated in Fig.~\ref{fig:model}. Our framework is hence trained using an anticipation 
loss, a recognition loss, and a cycle consistency loss. In the following sections, we describe in detail these modules.

\subsection{Sequence-to-Sequence Model}
The sequence-to-sequence model maps the sequence of observed frames to a sequence of future activities and their duration. 
For this model, we use a recurrent encoder-decoder architecture based on gated recurrent units (GRUs). 

\noindent\textbf{Recurrent Encoder.}
The purpose of the recurrent encoder is to encode the observed frames 
in a single vector which will be used to decode the future activities. 
Given the input features $X$, the recurrent encoder passes these features through a single layer with a gated 
recurrent unit (GRU)
\begin{equation}
h^e_t = GRU(x_{t}, \ h^e_{t-1}), 
\end{equation}
where $x_{t}$ is the input feature for frame $t$, and $h^e_{t-1}$ is 
the hidden state at the previous time step. 
The hidden state at the last time step $h^e_{t_o}$ encodes the observed frames and will be 
used as an initial state for the decoder.

\noindent\textbf{Recurrent Decoder.}
Given the output of the encoder $h^e_{t_o}$, the recurrent decoder predicts the future 
activities and their relative duration. The decoder also consists of a single layer with a 
gated recurrent unit (GRU). The hidden state at each step is updated using the GRU update rules
\begin{equation} \label{eqn:recurrent_decoder}
h^d_m = GRU(A_{m-1}, \ h^d_{m-1}),
\end{equation}
where the input at each time step $A_{m-1}$ is the predicted activity label for the previous step. 
At training time, the ground-truth label is used as input. Whereas the predicted label is used during 
inference time. For the first time step, a special start-of-sequence (SOS) symbol is used as input. 
Given the hidden state $h^d_m$ at each time step, the future activity label $A_m$ and its relative 
duration $\ell_m$ are predicted using a fully connected layer,~\ie
\begin{equation}
\tilde{A}_m = W_{A}h^d_m + b_{A},
\end{equation}
\begin{equation}
\hat{\ell}_m = W_{\ell}h^d_m + b_{\ell},
\end{equation}
where $\tilde{A}_m$ is the predicted logits for the future activity, and $\hat{\ell}_m$ is the predicted duration 
in log space. To get the relative duration, we apply softmax over the time steps
\begin{equation}
\tilde{\ell}_m = \frac{e^{\hat{\ell}_m}}{\sum_k e^{\hat{\ell}_k}}.
\end{equation}
The decoder keeps predicting future activity labels and the corresponding duration until a special 
end-of-sequence (EOS) symbol is predicted.
As a loss function, we use a combination of a cross entropy loss for the activity label and a mean 
squared error (MSE) for the predicted relative duration
\begin{align}
\mathcal{L}_{A} &= \mathcal{L}_{CE} + \mathcal{L}_{MSE}, \\
\mathcal{L}_{CE} &= \frac{1}{N-n}\sum_{m=n+1}^{N} -log(\tilde{a}_{m,A}), \\
\mathcal{L}_{MSE} &= \frac{1}{N-n}\sum_{m=n+1}^{N} (\tilde{\ell}_m - \ell_m)^2,
\end{align}
where $\mathcal{L}_{A}$ is the anticipation loss, $\tilde{a}_{m,A}$ is the the predicted probability 
for the ground truth activity label at step $m$, $n$ is the number of observed action segments 
and $N$ is the total number of action segments in the video. 

Since the input to the encoder are frame-wise features which might be very long, the output of the 
encoder $h^e_{t_o}$ might not be able to capture all the relevant information. To alleviate this problem 
we combine the decoder with an attention mechanism using a multi-head attention module~\cite{vaswani2017attention}. 
Additional details are given in the supplementary material.

\subsection{Cycle Consistency}
Since predicting the future from the past and the past from the future should be consistent, we propose an 
additional cycle consistency loss. 
Given the predicted future activities, we want to predict the past activities and their duration. 
This requires the predicted future to be good enough to predict the past. The cycle consistency loss 
has two benefits. First, it verifies if, for the 
predicted future actions, the actions that have been previously observed are plausible. 
Second, it encourages the recurrent decoder to keep the most important information of 
the observed sequence until the end, instead of storing only the information of the previous anticipated activity. 
In this way, a wrong prediction does not necessary propagate since the observed sequence is kept in memory.

The cycle consistency module is similar to the recurrent decoder and consists of a single layer with GRU, however, 
it predicts the past instead of the future. The hidden state of this GRU is initialized with the last hidden 
state of the recurrent decoder and at each step the hidden state is updated as follows
\begin{equation}
h^{cyc}_m = GRU(A_{m-1}, \ h^{cyc}_{m-1}).
\end{equation}
Given the hidden state $h^{cyc}_m$ at step $m$, an activity label of the observations and its relative duration 
are predicted using a fully connected layer. The loss function is also similar to the recurrent decoder
\begin{align}
\mathcal{L}_{cyc} &= \mathcal{L}_{CE} + \mathcal{L}_{MSE}, \\
\mathcal{L}_{CE} &= \frac{1}{n}\sum_{m=1}^{n} -log(\tilde{a}_{m,A}), \\
\mathcal{L}_{MSE} &= \frac{1}{n}\sum_{m=1}^{n} (\tilde{\ell}_m - \ell_m)^2,
\end{align}
where $\mathcal{L}_{cyc}$ is the cycle consistency loss, $\mathcal{L}_{CE}$ and $\mathcal{L}_{MSE}$ 
are the cross entropy loss and MSE loss applied on the past activity labels and their relative duration.

While the mapping from past to future can be multi-modal, this does not limit 
the applicability of the cycle consistency module. Since the cycle consistency 
module is conditioned on the predicted future, no matter what mode is predicted, 
the cycle consistency makes sure it is plausible. This also applies to the inverse 
mapping. As there is a path from the observed frames to the cycle consistency 
module through the sequence-to-sequence model, there is no ambiguity in which 
past activities have been observed.


\subsection{Recognition Module}
In the sequence-to-sequence model, the input of the recurrent encoder are the frame-wise features. However, directly 
passing the features to the encoder is sub-optimal as the encoder might struggle to capture all the relevant information 
for anticipating the future activities. As past activities provide a strong signal for anticipating future activities, 
we use a recognition module that learns discriminative features of the observed frames. These features will then serve as 
an input for the sequence-to-sequence model to anticipate the future activities. Given the success 
of temporal convolutional networks (TCNs) in analyzing activities in videos~\cite{Lea2017temporal,abufarha2019tcn}, 
we use a similar network for our recognition module. Besides being a strong model for analyzing videos, TCNs are fully 
differentiable and can be integrated in our framework without preventing end-to-end training.
For our module, we use a TCN similar to the one proposed in~\cite{abufarha2019tcn}. 
The TCN consists of several layers of dilated 1D convolutions where the dilation factor is doubled at each layer. 
The operations at each layer can be formally 
described as follows
\begin{align}
& \hat{F}_l = ReLU(W_1 * F_{l-1} + b_1), \\
& F_l = F_{l-1} + W_2 * \hat{F}_l + b_2, 
\end{align}
where $F_l$ is the output of layer $l$, $*$ is the convolution operator, 
$W_1 \in \mathbb{R}^{3 \times K \times K}$ are the weights of the dilated convolutional 
filters with kernel size 3 and $K$ is the number of the filters, 
$W_2 \in \mathbb{R}^{1 \times K \times K}$ are the weights of a $1 \times 1$ convolution, 
and $b_1, b_2 \in \mathbb{R}^{K}$ are bias vectors. The input of the first layer $F_0$ is obtained 
by applying a $1 \times 1$ convolution over the input features $X$. 

The output of the last dilated convolutional layer serves as input to the subsequent modules. To 
make sure that these features are discriminative enough, we add a classification layer that 
predicts the action label at each observed frame 
\begin{equation}
\tilde{Y}_t = Softmax(Wf_{L,t} + b), 
\end{equation}
where $\tilde{Y}_t$ contains the class probabilities at time $t$, $f_{L,t} \in \mathbb{R}^{K}$ is the output 
of the last dilated convolutional layer at time $t$, $W \in \mathbb{R}^{C \times K}$ and 
$b \in \mathbb{R}^{C}$ are the weights and bias for the $1 \times 1$ convolutional layer, 
where $C$ is the number of action classes.
To train this module we use a cross entropy loss
\begin{equation}
\mathcal{L}_{R} = \frac{1}{t_o}\sum_{t=1}^{t_o} -log(\tilde{y}_{t,c}), 
\end{equation}
where $\tilde{y}_{t,c}$ is the predicted probability for the ground truth 
label $c$ at time $t$, and $t_o$ is the number of observed frames.

\subsection{Loss Function}
To train our framework, we sum up all the three mentioned losses
\begin{equation}
\mathcal{L} = \mathcal{L}_{A} + \mathcal{L}_{R} + \mathcal{L}_{cyc}, 
\end{equation}
where $\mathcal{L}_{A}$ is the anticipation loss, $\mathcal{L}_{R}$ is the recognition 
loss, and $\mathcal{L}_{cyc}$ is the cycle consistency loss.


\section{Experiments}

We evaluate the proposed model on two datasets: the Breakfast dataset~\cite{kuehne2014language} 
and 50Salads~\cite{stein2013combining}. In all experiments, we report the average of three runs. 

The \textbf{Breakfast} dataset is a collection of $1,712$ videos with overall $66.7$ hours and 
roughly $3.6$ million frames. 
Each video belongs to one out of ten breakfast related activities, such as make tea or pancakes. 
The video frames are annotated with fine-grained action labels like \textit{pour milk} or \textit{fry egg}. Overall, 
there are $48$ different actions. On average, each video contains $6$ action instances and is $2.3$ minutes 
long. For evaluation, we use the standard 4 splits as proposed in~\cite{kuehne2014language} and report the average.

The \textbf{50Salads} dataset contains 50 videos showing people preparing different kinds of salad. 
These videos are relatively long with an average of $6.4$ minutes and $20$ action instances per video. 
The video frames are annotated with $17$ 
fine-grained action labels like \textit{cut tomato} or \textit{peel cucumber}. For evaluation, we use 
five-fold cross-validation and report the average as in~\cite{stein2013combining}. 

We follow the state-of-the-art evaluation protocol and report the mean over classes (MoC) accuracy for different observation/prediction 
percentages~\cite{abufarha2018when,ke2019time}.

\paragraph{Implementation Details.}
For the recognition module, we used a TCN with 10 layers and 64 filters in each layer. The number of units in 
the GRU cells is set to 512. For each training video, we generate two training examples with $20\%$ 
and $30\%$ observation percentage. The prediction percentage is always set to $50\%$. 
All the models are trained for 80 epochs using Adam optimizer~\cite{kingma2015adam}. We set the learning rate to $0.001$ and reduce it every $20$ epochs with a factor of $0.8$. 
For both datasets, we extract I3D~\cite{carreira2017quo} features for the video frames using both RGB and flow streams and sub-sample them at 5 frames per second.

\begin{table}[tb]\centering \setlength{\tabcolsep}{6pt}
\caption{Ablation study on the Breakfast dataset. Numbers represent mean over classes (MoC) accuracy.}
\resizebox{.97\columnwidth}{!}{%
\begin{tabular}{@{}lccccccccc@{}}\toprule
Observation \% & \multicolumn{4}{c}{$20\%$} &   & \multicolumn{4}{c}{$30\%$} \\
\cmidrule{2-5} \cmidrule{7-10} 
Prediction \% & $10\%$ & $20\%$ & $30\%$ & $50\%$ && $10\%$ & $20\%$ & $30\%$ & $50\%$ \\ \midrule
S2S & 
23.22       &       20.92       &       20.10      &       20.05      &&
26.43       &       24.38       &       24.13      &       23.38 \\

S2S + TCN  & 
14.52       &       13.55       &       13.15      &       12.70      &&
14.83       &       14.11       &       13.61      &       13.01 \\

S2S + TCN + $\mathcal{L}_R$ & 
24.72       &       22.43       &       21.70      &       21.77      &&
28.35       &       26.29       &       25.01      &       24.47 \\

S2S + TCN + $\mathcal{L}_R$ + $\mathcal{L}_{cyc}$  & 
25.16       &       22.73       &       22.22      & \textbf{22.01}  &&
28.07       &       26.25       &       25.12      &       24.81 \\

S2S + TCN + $\mathcal{L}_R$ + $\mathcal{L}_{cyc}$ + attn.      & 
\textbf{25.88} & \textbf{23.42} & \textbf{22.42} &         21.54   &&
\textbf{29.66} & \textbf{27.37} & \textbf{25.58} & \textbf{25.20} \\
\bottomrule
\end{tabular}
\label{tab:ablation_exp}
}
\end{table}

\subsection{Ablation Analysis}
In this section, we analyze the impact of the different modules in our framework on the anticipation performance. 
This analysis is conducted on the Breakfast dataset and the results are shown in Table~\ref{tab:ablation_exp}. 
Additional ablation experiments to study the impact of the input to the recurrent encoder and decoder are provided in the supplementary material.

\paragraph{Impact of the recognition module:} The recognition module consists of two parts: a TCN and a recognition loss $\mathcal{L}_{R}$. Starting from only the sequence-to-sequence module (S2S), we can achieve 
a good accuracy in the range $20\%-27\%$. By combining the sequence-to-sequence module with the recognition 
module (S2S + TCN + $\mathcal{L}_R$), we gain an improvement of $1\%-2\%$ for each observation-prediction percentage. 
This indicates that recognition helps the anticipation task. We also evaluate the performance when the TCN is 
combined with the sequence-to-sequence module 
without the recognition loss (S2S + TCN). As shown in Table~\ref{tab:ablation_exp}, the results are worse than using 
only the sequence-to-sequence module if we do not apply the recognition loss. This can be explained by the 
structure of the network. The recurrent encoder maps the features extracted by the TCN from all frames to 
a single vector. Without additional loss for the recognition module, the gradient vanishes and the parameters 
of the TCN are not well estimated. Nevertheless, by just applying the recognition loss, we get enough 
supervisory signal to train the TCN and improve the overall anticipation accuracy. This also highlights that 
the improvements from the recognition module are due to the additional recognition task and not because 
of having more parameters.

\begin{figure}[t]
\centering
\begin{tabular}{c@{\hskip .1in}c}
\includegraphics[trim={.3cm 3.5cm 2.3cm 1.5cm},clip,width=.48\textwidth]{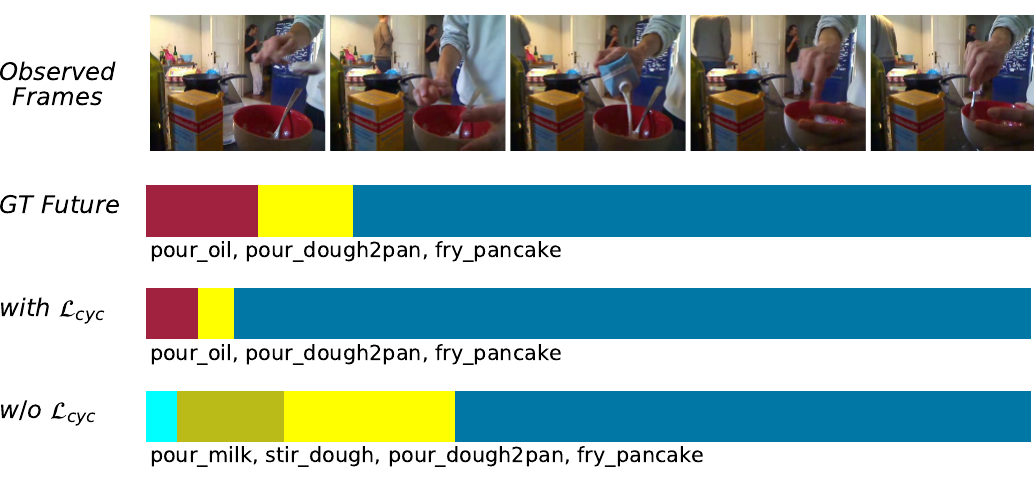} 
& 
\includegraphics[trim={.3cm 3.5cm 2.3cm 1.5cm},clip,width=.48\textwidth]{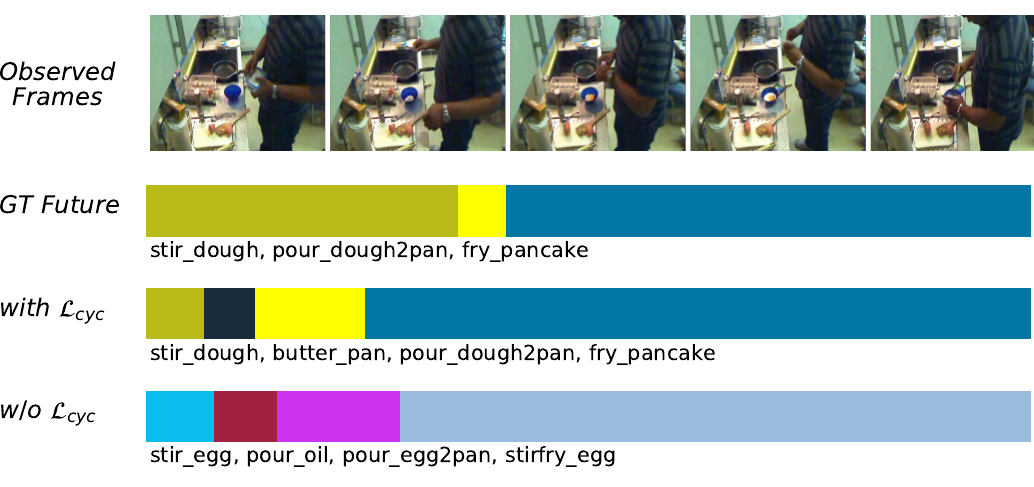}
\\
(a) &(b)\\
\end{tabular}
\vspace{-2mm}
\caption{Impact of the cycle consistency loss. Cycle consistency verifies if, for the predicted future
actions, all required actions have been done before and no essential action is missing (a), 
and it further encourages the decoder to keep the important information from the observations 
until the end, which results in better predictions (b).}
\label{fig:cc_q_res}
\end{figure}

\begin{table}[tb]\centering  \setlength{\tabcolsep}{6pt}
\caption{Comparison between the two-step approach and ours on the Breakfast dataset. Numbers represent mean over classes (MoC) accuracy.}
\resizebox{.97\columnwidth}{!}{%
\begin{tabular}{@{}@{\hskip .35in}c@{\hskip .55in}ccccccccc@{}}\toprule
Observation \% & \multicolumn{4}{c}{$20\%$} &   & \multicolumn{4}{c}{$30\%$} \\
\cmidrule{2-5} \cmidrule{7-10} 
Prediction \% & $10\%$ & $20\%$ & $30\%$ & $50\%$ && $10\%$ & $20\%$ & $30\%$ & $50\%$ \\ \midrule
Two-Step & 
23.58       &       21.90       &       20.80      &       20.18      &&
27.80       &       25.21       &       23.32      &       22.96 \\

Ours      & 
\textbf{25.88} & \textbf{23.42} & \textbf{22.42} & \textbf{21.54}   &&
\textbf{29.66} & \textbf{27.37} & \textbf{25.58} & \textbf{25.20} \\
\bottomrule
\end{tabular}
}
\label{tab:two_step_vs_ours}
\vspace{-4mm}
\end{table}

\paragraph{Impact of the cycle consistency loss:} The cycle consistency module predicts the past 
activities from the predicted future. The intuition is that to be able to predict the past activities, 
the predicted future activities have to be correct. As shown in Table~\ref{tab:ablation_exp}, using 
the cycle consistency loss gives an additional improvement on the anticipation accuracy. 
The cycle consistency loss verifies if, for the predicted future actions, all required actions have been 
done before and no essential action is missing. For example in Fig.~\ref{fig:cc_q_res} (a), the model 
observes \textit{spoon flour}, \textit{crack egg}, \textit{pour milk}, and \textit{stir dough}. Without the cycle consistency the network did 
not predict the action \textit{pour oil}, which is required to \textit{fry pancake}. By using cycle consistency this issue 
is resolved. Cycle consistency also forces the decoder to remember all observed activities. As illustrated 
in Fig.~\ref{fig:cc_q_res} (b), the model observes \textit{spoon flour}, \textit{crack egg}, \textit{pour milk}, \textit{butter pan}, and \textit{stir dough}. Without the cycle consistency, the network predicts the action \textit{stirfry egg}, which would have been 
plausible if \textit{spoon flour} and \textit{pour milk} were not part of the observations. Since the cycle consistency encourages the decoder to use all observed actions for anticipation, the activity \textit{fry pancake} is correctly anticipated.

\paragraph{Impact of the attention module:} Finally, using the full model by combining the recurrent decoder with 
the multi-head attention module further improves the results by roughly $1\%$. As shown in Table~\ref{tab:ablation_exp}, 
the gain from using the attention module is higher when the observation percentage is $30\%$. This is mainly because 
of the encoder module. Given the observed frames, the encoder tries to encode them in a single vector. This means that 
the encoder has to throw away more information from the long sequences compared to shorter sequences. In this case, 
the attention module can help in capturing some of the lost information in the encoder output by attending on the 
relevant information in the observations.

\subsection{End-to-End vs. Two-Step Approach}

To illustrate the benefits of end-to-end learning over two-step approaches, we compare our framework with its two-step 
counterpart. For this comparison, we first train the recognition module from our framework and then fix the 
weights of the TCN and train the remaining components of our model with the anticipation loss and the cycle 
consistency loss. Table~\ref{tab:two_step_vs_ours} shows the results of our framework compared to the two-step 
approach on the Breakfast dataset with different observation and prediction percentages. As shown in the table, 
our framework outperforms the two-step approach with a large margin of up to $2.3\%$. This highlights the benefits 
of end-to-end approaches where the model can capture the relevant information in the observed frames to anticipate the future. 
On the contrary, two-step approaches can only utilize the label information of the observed frames that are not optimized 
for the anticipation task which is sub-optimal.


\begin{table}[tb]\centering \scriptsize \setlength{\tabcolsep}{6pt}
\caption{Comparison with the state-of-the-art. Numbers represent MoC accuracy.}
\resizebox{.97\columnwidth}{!}{%
\begin{tabular}{@{}cccccccccc@{}}\toprule
Observation \% & \multicolumn{4}{c}{$20\%$} &   & \multicolumn{4}{c}{$30\%$} \\
\cmidrule{2-5} \cmidrule{7-10} 
Prediction \% & $10\%$ & $20\%$ & $30\%$ & $50\%$ && $10\%$ & $20\%$ & $30\%$ & $50\%$ \\ \midrule
\multicolumn{10}{l}{\textbf{Breakfast}} \\
\midrule 
RNN model~\cite{abufarha2018when} & 
18.11       &      17.20  &   15.94  &  15.81  &&
21.64       &      20.02   &  19.73  &  19.21 \\
CNN model~\cite{abufarha2018when} & 
17.90	   &	16.35		&	15.37	&	14.54		&&
22.44	   &	20.12  	    &	19.69	&	18.76	 \\
RNN~\cite{abufarha2018when} + TCN& 
05.93	   &	05.68		&	05.52	&	05.11		&&
08.87	   &	08.90  	    &	07.62	&	07.69	 \\
CNN~\cite{abufarha2018when} + TCN& 
09.85	   &	09.17		&	09.06	&	08.87		&&
17.59	   &	17.13  	    &	16.13	&	14.42	 \\
UAAA (mode)~\cite{abufarha2019uncertainty} & 
16.71      &       15.40       &       14.47      &       14.20      &&
20.73      &       18.27       &       18.42      &       16.86\\
Time-Cond.~\cite{ke2019time} & 
18.41      &       17.21       &       16.42      &       15.84      &&
22.75      &       20.44       &       19.64      &       19.75\\
Ours      & 
\textbf{25.88} & \textbf{23.42} & \textbf{22.42} & \textbf{21.54}   &&
\textbf{29.66} & \textbf{27.37} & \textbf{25.58} & \textbf{25.20} \\
\bottomrule 
\multicolumn{10}{l}{\textbf{50Salads}} \\
\midrule 
RNN model~\cite{abufarha2018when} & 
30.06	&	25.43       	&	18.74       	&	13.49	&&
30.77	&	17.19			&	14.79			&	09.77 \\
CNN model~\cite{abufarha2018when} & 
21.24			&	19.03		  &	    15.98		  & 	09.87	&&
29.14			&	20.14	      &	    17.46	      &	    10.86	 \\
RNN~\cite{abufarha2018when} + TCN& 
32.31	   &	25.51		&	19.10	&	14.15		&&
26.14	   &	17.69  	    &	16.33	&	12.97	 \\
CNN~\cite{abufarha2018when} + TCN& 
16.02	   &	14.68		&	12.09	&	09.89		&&
19.23	   &	14.68  	    &	13.18	&	11.20	 \\
UAAA (mode)~\cite{abufarha2019uncertainty} & 
24.86      &       22.37       &       19.88      &       12.82      &&
29.10      &       20.50       &       15.28      &       12.31\\
Time-Cond.~\cite{ke2019time} & 
32.51      &       27.61       &       21.26   &   \textbf{15.99}      &&
\textbf{35.12} & \textbf{27.05} &       \textbf{22.05}      &       15.59\\
Ours      & 
\textbf{34.76} & \textbf{28.41} & \textbf{21.82}    &    15.25      &&
34.39       &       23.70       &       18.95      &       \textbf{15.89} \\
\bottomrule
\end{tabular}
}
\label{tab:SoA}
\end{table}

\subsection{Comparison with the State-of-the-Art}
In this section, we compare our framework with the state-of-the-art methods on both the Breakfast dataset 
and 50Salads. We follow the same protocol and report results for different observation and prediction percentages. 
Table~\ref{tab:SoA} shows the results on both datasets. 
All the previous approaches follow the two-step approach by inferring the action labels 
of the observed frames first and then use these labels to anticipate the future activities. As shown in 
Table~\ref{tab:SoA}, our framework outperforms all the state-of-the-art methods by a large margin of 
roughly $5\%-8\%$ for each observation-prediction percentage pair on the Breakfast dataset. An interesting observation is that all the 
previous approaches achieve comparable results despite the fact that they are using different network architectures 
based on RNNs~\cite{abufarha2019uncertainty,abufarha2018when}, CNNs~\cite{abufarha2018when}, or even temporal convolution~\cite{ke2019time}. 
On the contrary, our framework clearly outperforms these approaches with the advantage that it was trained 
in an end-to-end fashion.

For 50Salads, our model outperforms the state-of-the-art in $50\%$ of the cases. This is mainly 
because 50Salads is a small dataset. Since our model is trained end-to-end, it requires more data to show 
the benefits over two-step approaches.

\begin{figure}[t]
\centering
\begin{tabular}{c@{\hskip .1in}c}
\includegraphics[trim={.4cm 2.5cm 2.3cm 1.5cm},clip,width=.48\textwidth]{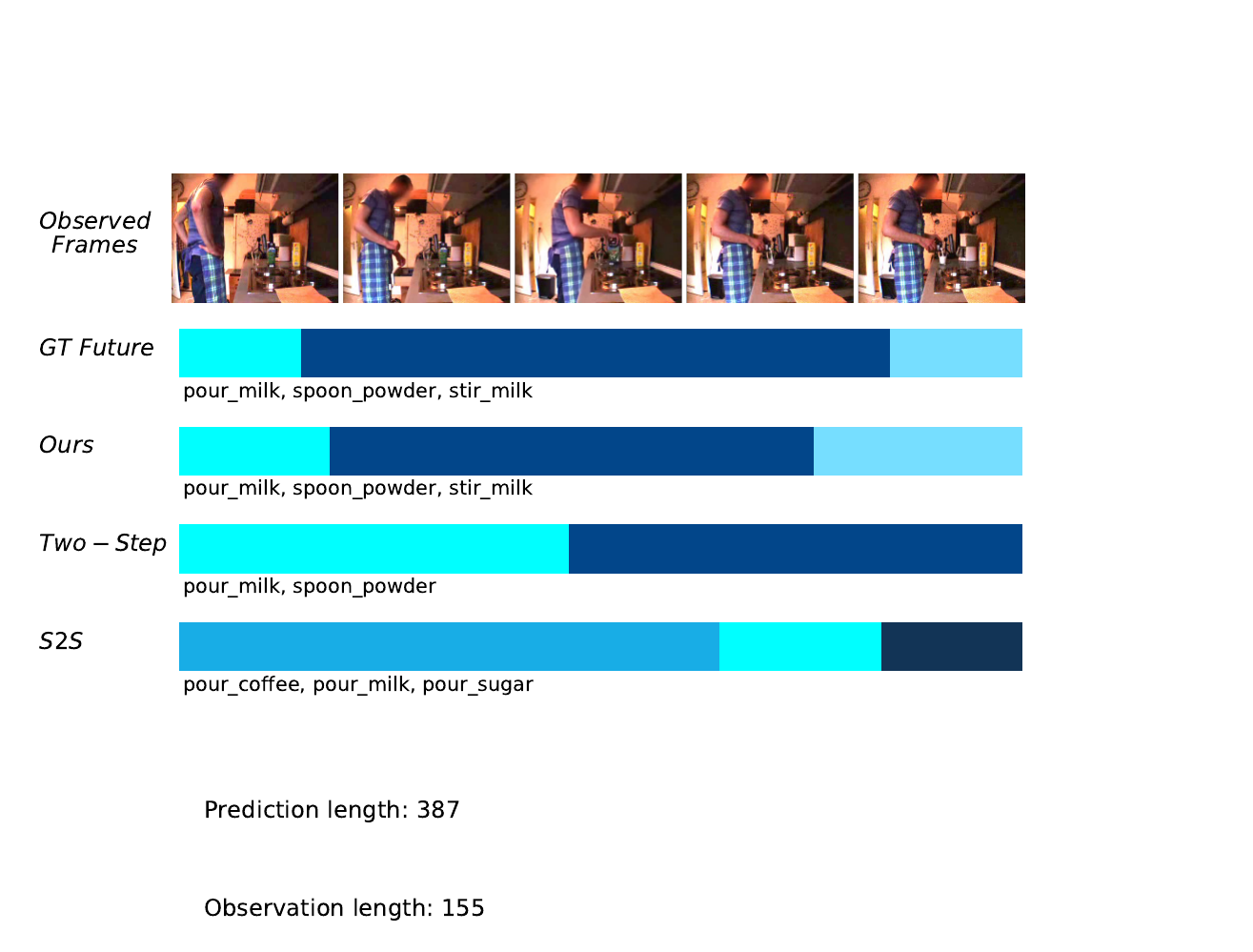} 
& 
\includegraphics[trim={.4cm 2.5cm 2.3cm 1cm},clip,width=.48\textwidth]{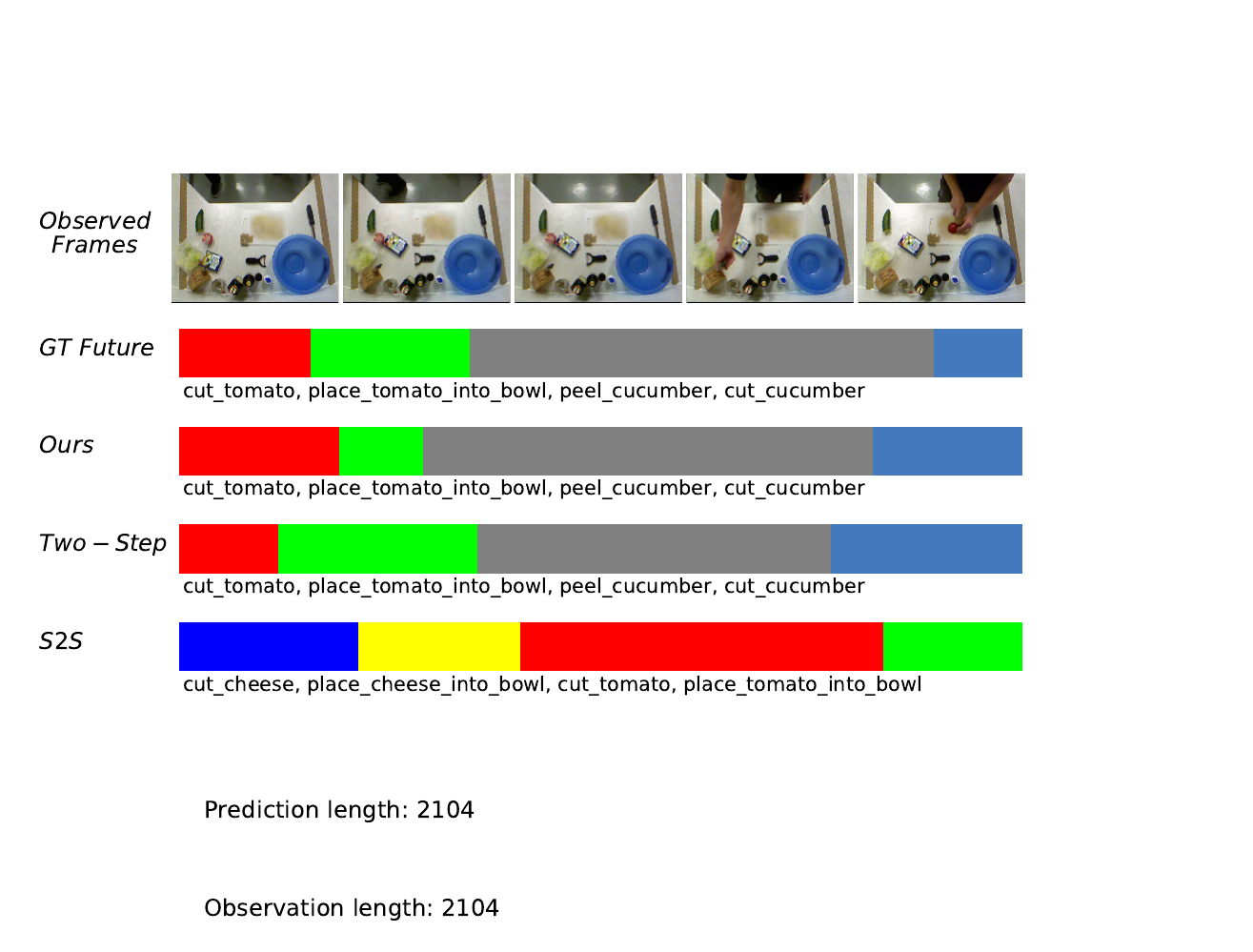} 
\\
(a) &(b)\\
\end{tabular}
\vspace{-2mm}
\caption{Qualitative results for anticipating future activities. (a) An example from the 
Breakfast dataset for the case of observing $20\%$ of the video and predicting the activities 
in the following $50\%$. (b) An example from the 50Salads dataset for the case of observing $20\%$ 
of the video and predicting the activities in the following $20\%$.}
\label{fig:q_res}
\vspace{-2mm}
\end{figure} 

Since the state-of-the-art methods like~\cite{abufarha2018when} use an RNN-HMM model~\cite{richard2017weakly} for recognition, we also report 
the results of~\cite{abufarha2018when} with our TCN as a recognition model. 
The results are shown in Table~\ref{tab:SoA} 
(RNN~\cite{abufarha2018when} + TCN and CNN~\cite{abufarha2018when} + TCN). Our model outperforms 
these methods even when they are combined with TCN. This highlights that the improvements in our model are not only due 
to the TCN, but mainly because of the joint optimization of all modules for the anticipation task in 
an end-to-end fashion and the introduced cycle consistency loss.

Qualitative results for our model on both datasets are illustrated in Fig.~\ref{fig:q_res}. As shown in the figure, 
our model can generate accurate predictions of the future activities and their duration. We also show the 
results of the sequence-to-sequence (S2S) and the two-step baselines. Our model anticipates 
the activities better. 


\section{Conclusion}
In this paper, we introduced a model for anticipating future activities from a partially 
observed video. In contrast to the state-of-the-art methods which rely on the action labels 
of the observations, our model directly predicts the future activities from the observed frames. 
We train the proposed model in an end-to-end fashion and show a superior performance compared to 
the previous approaches. Additionally, we introduced a cycle consistency loss for the anticipation 
task which further boosts the performance. Our framework achieves state-of-the-art results on two 
publicly available datasets.
\\

\footnotesize{
\noindent\textbf{Acknowledgments:}
The work has been funded by the Deutsche Forschungsgemeinschaft (DFG, German Research Foundation) – GA 1927/4-1 (FOR 2535 Anticipating Human Behavior) and the ERC Starting Grant ARCA (677650).}

	\bibliographystyle{splncs03}
	\bibliography{egbib}

	\pagestyle{headings}
	\mainmatter

	\def\GCPR20SubNumber{22}

	\title{Long-Term Anticipation of Activities with Cycle Consistency \\ (Supplementary Material)}

	\titlerunning{Long-Term Anticipation of Activities with Cycle Consistency.}
	\authorrunning{Abu Farha \etal.}
	\author{Yazan Abu Farha$^{1}$ \and Qiuhong Ke$^{2}$ \and Bernt Schiele$^{3}$ \and Juergen Gall$^{1}$}
	\institute{$^{1}$University of Bonn, Germany \\
	           $^{2}$The University of Melbourne, Australia \\
	           $^{3}$MPI Informatics, Germany}

	\maketitle

We provide more details about the attention module that we use for 
the sequence-to-sequence model. We further provide more ablation experiments to study the impact of the input to the recurrent encoder and decoder. 

\section{The Attention Module}

For the sequence-to-sequence model, we use a recurrent encoder-decoder architecture based on 
gated recurrent units (GRUs). The encoder encodes the observed frames in a single vector which 
will be used to decode the future activities.
Since the input to the encoder are frame-wise features which might be very long, the output of the 
encoder $h^e_{t_o}$ might not be able to capture all the relevant information. To alleviate this problem 
we combine the decoder with an attention mechanism using a multi-head attention module.

Given the current hidden state of the decoder $h^d_m$ and the output of the encoder at each step 
$h^e_{1:t_o}=(h^e_1, \dots, h^e_{t_o})$, the attention output for a single head is computed by first 
generating the query (q), the keys (K) and values (V) as described in the following
\begin{align}
q &= W_q h^d_m, \\
K &= W_k h^e_{1:t_o},  \\
V &= W_v h^e_{1:t_o},
\end{align}
where $q \in \mathbb{R}^{d_A}$ is a column vector, $K,\ V \in \mathbb{R}^{d_A \times t_o}$ and 
$d_A$ is set to be one-eighth of the hidden state size of the GRU.
We normalize these tensors by the $\left\Vert . \right\Vert_2$ norm and then compute the output 
using a weighted sum of the values 
\begin{align}
O_h &= Softmax(q^T K) V^T, 
\end{align}
where $O_h \in \mathbb{R}^{d_A}$ is the output of a single head. We use $8$ heads 
and concatenate the output of these heads, apply a linear transformation and then concatenate the output with 
the input of the decoder at each step.

\section{Frame-wise Features vs. Frame-wise Labels}

In our model, we use the output features of the recognition module as input to the subsequent modules. 
Since the recognition module also predicts frame-wise class probabilities of the observed frames, another option 
would be to pass these probabilities instead of the features. In this section, we provide a comparison 
between these two variants. As shown in Table~\ref{tab:labels_vs_features}, passing frame-wise class probabilities 
gives a comparable performance to frame-wise features. Nevertheless, using the features is slightly better. This 
is expected as the features contain much richer information that can be utilized by the subsequent modules. We 
also tried concatenating the frame-wise features and the frame-wise class probabilities and pass the concatenated 
tensor to the sequence-to-sequence module. However, it did not improve the results.

\begin{table}[tb]\centering \setlength{\tabcolsep}{6pt}
\caption{Comparison between passing features vs. passing class probabilities of the observed frames 
from the recognition module to the subsequent modules on the Breakfast dataset. Numbers represent 
mean over classes (MoC) accuracy.}
\resizebox{.97\columnwidth}{!}{%
\begin{tabular}{@{}cccccccccc@{}}\toprule
Observation \% & \multicolumn{4}{c}{$20\%$} &   & \multicolumn{4}{c}{$30\%$} \\
\cmidrule{2-5} \cmidrule{7-10} 
Prediction \% & $10\%$ & $20\%$ & $30\%$ & $50\%$ && $10\%$ & $20\%$ & $30\%$ & $50\%$ \\ \midrule
Ours with class probabilities & 
25.53       &       23.39       &       22.50   & \textbf{21.70}      &&
28.95       &       27.36       &       25.06      &       24.32 \\

Ours with features     & 
\textbf{25.88} & \textbf{23.42} &         22.42  &         21.54   &&
\textbf{29.66} & \textbf{27.37} & \textbf{25.58} & \textbf{25.20} \\

Ours with features and prob. & 
25.15       &        23.41 &    \textbf{22.53}  &         21.64   &&
29.21       &        26.62 &         24.65  &         24.69  \\
\bottomrule
\end{tabular}
}
\label{tab:labels_vs_features}
\end{table}

\begin{table}[tb]\centering \setlength{\tabcolsep}{6pt}
\caption{Impact of passing the predicted duration at each step to the recurrent decoder on the Breakfast dataset. 
Numbers represent mean over classes (MoC) accuracy.}
\resizebox{.97\columnwidth}{!}{%
\begin{tabular}{@{}cccccccccc@{}}\toprule
Observation \% & \multicolumn{4}{c}{$20\%$} &   & \multicolumn{4}{c}{$30\%$} \\
\cmidrule{2-5} \cmidrule{7-10} 
Prediction \% & $10\%$ & $20\%$ & $30\%$ & $50\%$ && $10\%$ & $20\%$ & $30\%$ & $50\%$ \\ \midrule
Ours with duration input & 
25.21       &       20.94       &       19.61      &       20.38      &&
28.20       &       24.71       &       23.58      &       23.15 \\

Ours without duration input      & 
\textbf{25.88} & \textbf{23.42} & \textbf{22.42} & \textbf{21.54}   &&
\textbf{29.66} & \textbf{27.37} & \textbf{25.58} & \textbf{25.20} \\
\bottomrule
\end{tabular}
}
\label{tab:decoder_with_length_input}
\end{table}

\section{Impact of Passing the Duration to the Decoder}

The GRU in the recurrent decoder takes as input at each step the predicted activity label from the previous 
step as described in (\ref{eqn:recurrent_decoder}) in the main paper. In this section, we study the impact of passing the 
predicted duration to the decoder as well, \ie
\begin{equation} 
h^d_m = GRU([ A_{m-1}, \hat{\ell}_{m-1} ], \ h^d_{m-1}).
\end{equation}
As shown in Table~\ref{tab:decoder_with_length_input}, passing the duration to the decoder results in a degradation 
in performance. This is due to passing the duration before applying the softmax, which means that there might be 
huge variations in the duration value between different iterations. While passing the duration after applying the 
softmax is not feasible, passing only the activity labels provides enough information for the decoder. We also tried 
to predict the absolute duration. In this case, no EOS symbol is predicted and the decoder keeps generating 
future activities and their duration until the desired time horizon is predicted. However, the model performed very 
poorly in this setup. This is expected since predicting a stopping symbol is much easier than predicting the absolute 
duration of all activities.

\end{document}